\documentclass[runningheads]{llncs}
\usepackage{graphicx}
\usepackage{tabularx}
\usepackage{threeparttable}
\usepackage{subfig}
\usepackage{amsmath}
\usepackage{textcomp}
\usepackage{hyperref}
\usepackage{import}
\usepackage{graphics}
\usepackage{array}
\usepackage{booktabs}
\usepackage{caption}
\usepackage{xcolor}
\usepackage{enumitem}
\usepackage{tabularx}
\usepackage{multirow}
\usepackage{stfloats}
\usepackage{amssymb}
\usepackage{latexsym}
\usepackage{pbox}
\usepackage{multicol}
\usepackage{tikz}
\usepackage{float}
\usepackage[export]{adjustbox}
\usepackage[framemethod=TikZ]{mdframed}
\usepackage{cite}                                       
                                              
\begin{document}
\title{Document Visual Question Answering Challenge 2020}
%
%
\author{Minesh Mathew\inst{1} \and
Rubèn Tito\inst{2} \and
Dimosthenis Karatzas\inst{2} \and
R. Manmatha\inst{3} \and
C.V. Jawahar\inst{1}
}
\authorrunning{Mathew et al.}
%
\institute{CVIT, IIIT Hyderabad, India \and
Computer Vision Center, UAB, Spain \and
University of Massachusetts Amherst, USA\\
}
\maketitle              
\begin{abstract}
This paper presents results of Document Visual Question Answering Challenge organized as part of ``Text and Documents in the Deep Learning Era" workshop,  in CVPR 2020. The challenge introduces a new problem - Visual Question Answering on document images. The challenge comprised two tasks. 
The first task concerns with asking questions on a single document image.
On the other hand, the second task is set as a retrieval task where the question is posed over a collection of images. 
For the task 1 a new dataset is introduced comprising $50,000$ questions-answer(s) pairs defined over $12,767$ document images. For task 2 another  dataset has been created comprising $20$ questions over $14,362$ document images which share the same document template.
\end{abstract}
%
%
\section{Introduction}
Visual Question Answering (VQA) has attracted an intense research effort over the past few years, as one of the most important tasks at the frontier between vision and language. Notably, at the same time as reading systems research considered that the field of scene text understanding was mature enough to build scene text based VQA systems on, VQA researchers realised that the capacity of reading is actually important for any VQA agent. As a result, ST-VQA~\cite{stvqa_iccv} and TextVQA~\cite{singh2019towards} were introduced in parallel in 2019
In this year's Visual Question Answering workshop at CVPR 2020, three out of the five VQA challenges actually revolve around text~\cite{jayasundara2019textcaps, gurari2018vizwiz}.
The time seemed right for the introduction of a large scale scanned Document VQA task.

In this short paper we introduce the DocVQA dataset and the challenge organized as part of the ``Text and Documents in the Deep Learning Era" workshop at CVPR 2020. The paper offers a quick description of the dataset, the challenge and results of the submitted methods to date. The challenge is open for continuous submission at the Robust Reading Competition (RRC) portal~\footnote{\url{https://rrc.cvc.uab.es/?ch=17}}.

\begin{figure}[t]
\centering
\begin{minipage}{0.45\textwidth}%
\subfloat{
\includegraphics[width=\textwidth]{images/ljvl0226_1.png}
\label{fig:subfig1}} \\\begin{tabular}{ l } 
 \\
Who is the message for ?\\ 
What is the date of the message ?\\ \\ \\
 
\end{tabular}
\label{table:questions_task1}

\end{minipage}%
\qquad
\begin{minipage}{0.45\textwidth}%
\subfloat{
\includegraphics[width=\textwidth]{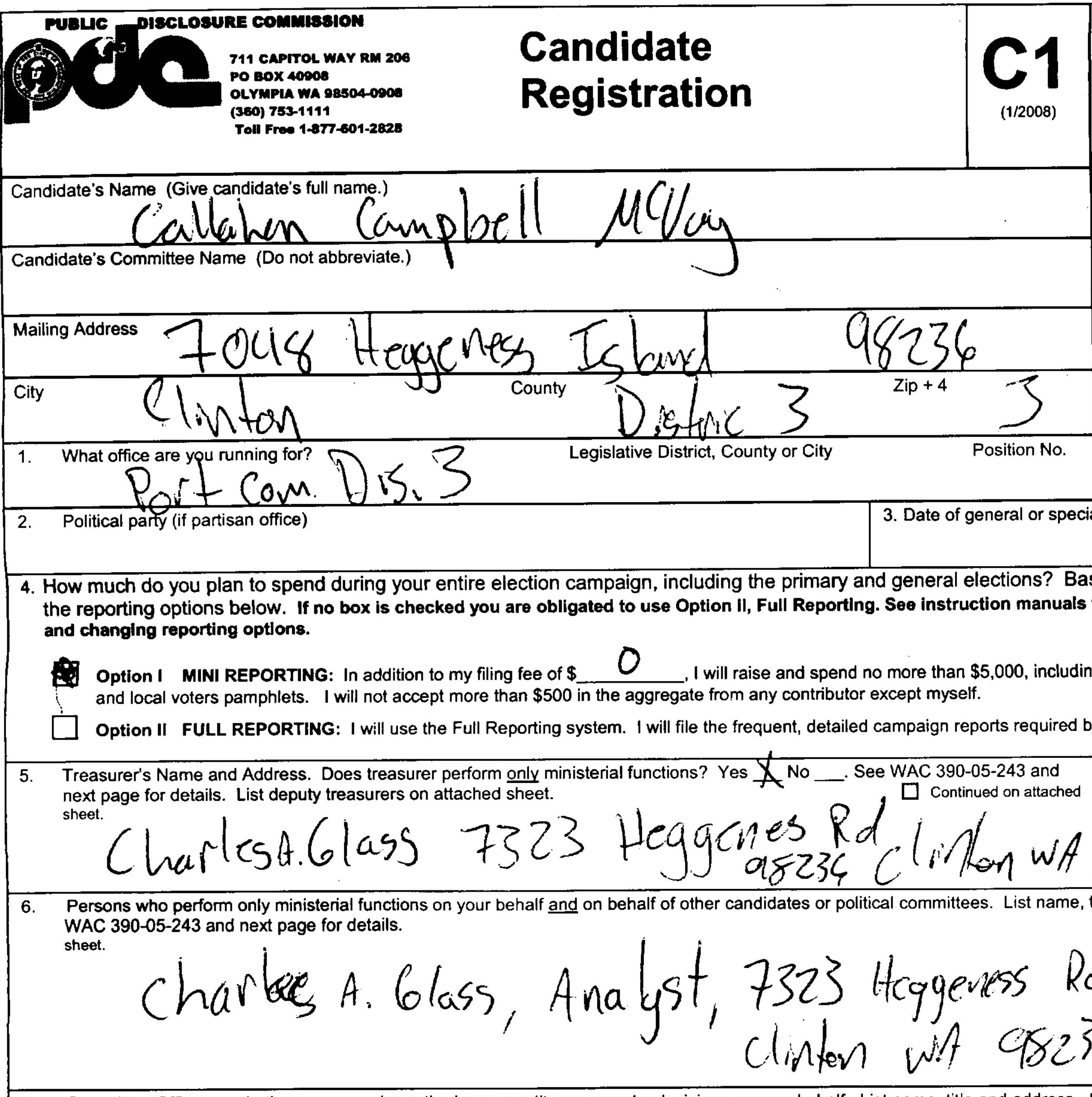}
\label{fig:subfig2}} \\

\begin{tabular}{ p{5.5cm} } 
\\

In which legislative counties did Gary L. Schoessler run for County Commissioner?
 
\end{tabular}
\label{table:questions_task2}
\end{minipage}
\caption{\textbf{(Left)} A sample document image and questions defined on it from the dataset for task 1. \textbf{(Right)} A sample document image from dataset for task 2 and a sample question posed on the whole task 2 document collection.}
\vspace{-0.5cm}
\label{fig:globfig}
\end{figure}
\section{Challenge}
The challenge ran between March and May 2020.
Ranking of submitted methods presented in this report reflect state of submissions at the closure of the official challenge period.

The challenge comprised two separate tasks.

\subsection{Task 1 - VQA on Document Images}
Task 1 of the challenge is similar to the typical VQA setting , i.e answer a question asked on an image; here a document image. Answer to the question is always text present in the image, or in other words it is an extractive QA task. The participants are required to submit their results on the test split of the dataset which comprise of 5188 questions defined on 1287 document images. For evaluation of the submissions, we use the same metric as ST-VQA challenge~\cite{st-vqa_challenge}, which is Average Normalized Levenshtein Similarity (ANLS).

\subsection{Task 2 - VQA on Document Image Collection}
In task 2 the question is posed over a collection of documents instead of a single image. Hence, the task requires one to retrieve the evidence as well output the answer.
For this first edition, we focused on the first part to rank the methods, and left the second as an optional response, which is nevertheless evaluated.
Performance of the methods for the retrieval part is scored by the Mean Average Precision (MAP), which is the standard metric in retrieval scenarios. It is important to note that we force positive evidences that have been equally scored by the methods, to be at the end of the ranking among those documents. This is to ensure that the ranking is consistent and do not depend on the default order, or the way the score is evaluated. Finally, in spite of the fact that answers score is not used to rank the methods, precision and recall are used to show the performance of the methods in this task.
%
%

%
\section{Datasets}
Dataset for task 1~\cite{mathew2020docvqa} comprises $50,000$ questions over $12,767$ document images. The images are pages extracted from 6071 scanned documents sourced from the industry documents library~\footnote{\url{https://www.industrydocuments.ucsf.edu/}}. We manually selected the documents from the library to ensure document variety.
Each question-answer(s) pair in the dataset is also qualified with a question type among 9 types that denote the type of analysis required to arrive at the answer. The 9 question types are figure, `form', `table/list', `layout', `running text', `photograph', `handwritten', `yes/no' and `other'. We also provide commercial OCR transcriptions of all the documents.

Dataset for task 2 consists of $14,362$ document images sharing the same document template --- US Candidate Registration form.
Among these images there are documents filled in handwriting like \autoref{fig:globfig} (right), while others are typewritten. 
The images were sourced from the Open Data portal of the Public Disclosure Commission (PDC) and over this collection a set of $20$ different questions is posed.

\section{Results}
\textbf{Task 1:} We received submissions from 9 different teams for task 1.
The final ranking is shown in Table~\ref{table:task1_ranking}.
Performance for the 9 submissions for different question types is shown in Figure~\ref{fig:qt_distribution}.

\begin{table}[ht]
\begin{minipage}[b]{0.35\linewidth}
\centering
\begin{tabularx}{\linewidth}{X c}
Method & ANLS  \\ \toprule 
PingAn-OneConnect-Gammalab-DQA & 0.85\\
Structural LM- v2 & 0.75\\
QA\_Base\_MRC\_1 & 0.74 \\ 
HyperDQA\_V4 &  0.69 \\ 
bert fulldata fintuned &  0.59 \\ 
Plain BERT QA &  0.35 \\ 
HDNet &  0.34 \\ 
CLOVA OCR &  0.33 \\ 
docVQAQV\_V0.1 &  0.30 \\  \bottomrule 
 
\end{tabularx}
\caption{\footnotesize Final ranking for task 1}
\vspace{-0.4cm}
\label{table:task1_ranking}
 
\end{minipage}\hfill
\begin{minipage}[b]{0.6\linewidth}
\centering
\includegraphics[width=\linewidth]{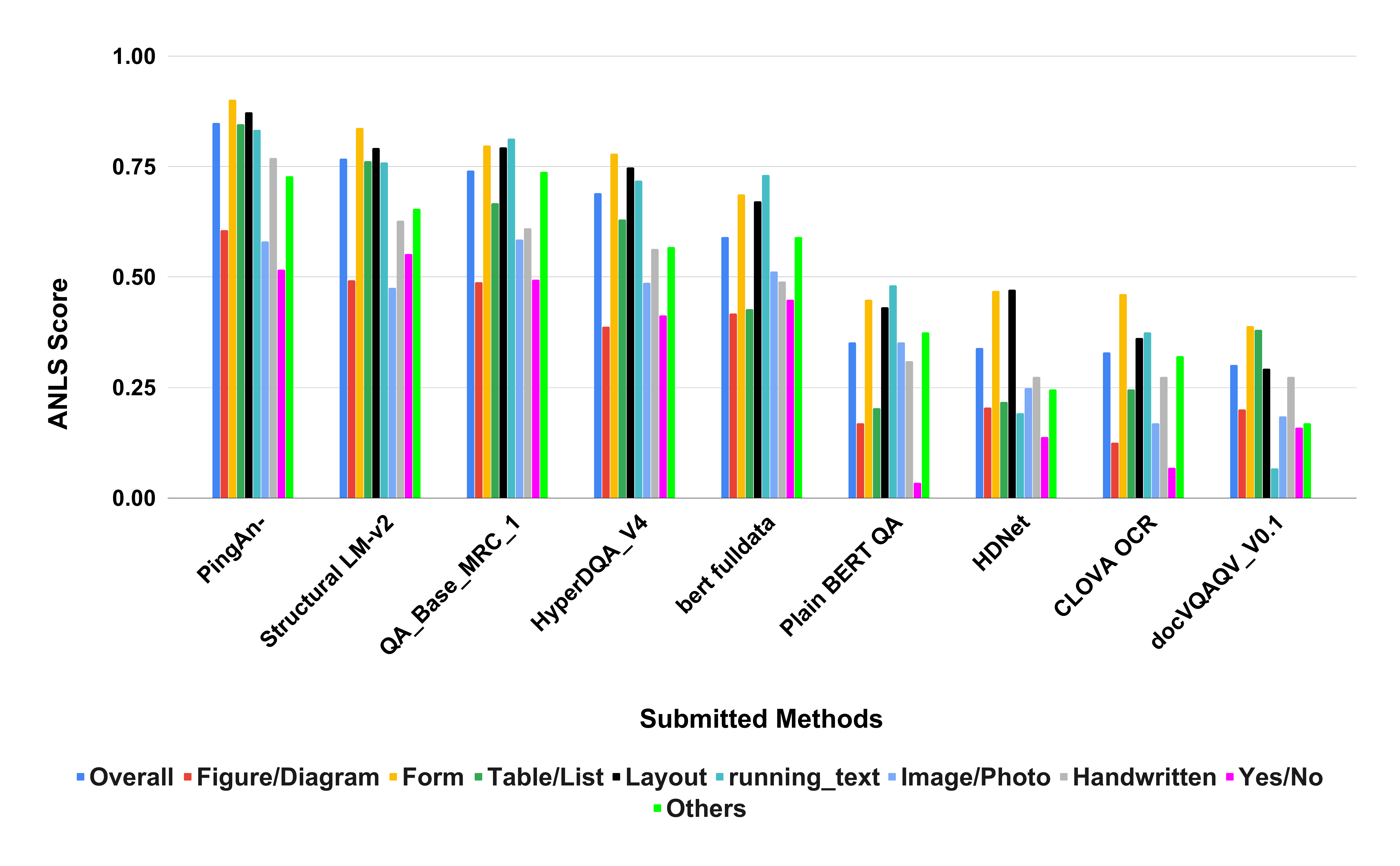} 
\captionof{figure}{\footnotesize Performance of the submitted methods for different question types in task 1.}
\vspace{-0.5cm}
\label{fig:qt_distribution}
\end{minipage}
\end{table}

\begin{table}[h]
\begin{minipage}[h]{0.45\linewidth}
\centering
    \begin{tabular}{lllcl}
           &   & Retrieval & \multicolumn{2}{c}{Answers} \\
    Method &   & MAP       & Precision      & Recall     \\ \midrule
    DQA    &   & 0.8090    & -              & -          \\
    DOCR   &   & 0.7915    & -              & -          \\ \bottomrule
    \end{tabular}
    \caption{\footnotesize Final ranking for task 2.}
    \vspace{-0.4cm}

   \label{tab:Task2_Results}

\end{minipage}\hfill
\begin{minipage}[c]{0.5\linewidth}
\centering
    \includegraphics[width=\linewidth]{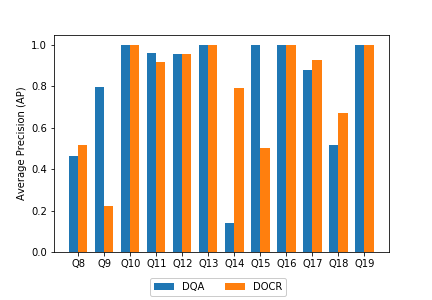}
    \captionof{figure}{\footnotesize Average precision of the submitted methods for each question in the test set.}
    \vspace{-0.7cm}

    \label{fig:task2_qap}
\end{minipage}
\end{table}
\noindent
\textbf{Task 2:} We received submissions from two different teams. The winning method is DQA from PingAn team, followed by a small margin by DOCR from iFLYTEK team (see \autoref{tab:Task2_Results}). None of the submitted methods provided the answers to the questions.

\section{Acknowledgements}
This work has been supported by Amazon through an AWS Machine Learning Research Award

\bibliographystyle{splncs04}
\footnotesize\bibliography{references}

\begin{thebibliography}{1}
\providecommand{\url}[1]{\texttt{#1}}
\providecommand{\urlprefix}{URL }
\providecommand{\doi}[1]{https://doi.org/#1}

\bibitem{st-vqa_challenge}
Biten, A.F., Tito, R., Mafla, A., G{\'{o}}mez, L., Rusi{\~{n}}ol, M., Mathew,
  M., Jawahar, C.V., Valveny, E., Karatzas, D.: {ICDAR} 2019 competition on
  scene text visual question answering. CoRR  \textbf{abs/1907.00490} (2019)

\bibitem{stvqa_iccv}
Biten, A.F., Tito, R., Mafla, A., Gomez, L., Rusinol, M., Valveny, E., Jawahar,
  C., Karatzas, D.: Scene text visual question answering. In: ICCV (2019)

\bibitem{gurari2018vizwiz}
Gurari, D., Li, Q., Stangl, A.J., Guo, A., Lin, C., Grauman, K., Luo, J.,
  Bigham, J.P.: Vizwiz grand challenge: Answering visual questions from blind
  people. In: CVPR. pp. 3608--3617 (2018)

\bibitem{jayasundara2019textcaps}
Jayasundara, V., Jayasekara, S., Jayasekara, H., Rajasegaran, J., Seneviratne,
  S., Rodrigo, R.: Textcaps: Handwritten character recognition with very small
  datasets. In: WACV. pp. 254--262. IEEE (2019)

\bibitem{mathew2020docvqa}
Mathew, M., Karatzas, D., Jawahar, C.: Docvqa: A dataset for vqa on document
  images. In: WACV (2021)

\bibitem{singh2019towards}
Singh, A., Natarajan, V., Shah, M., Jiang, Y., Chen, X., Batra, D., Parikh, D.,
  Rohrbach, M.: Towards vqa models that can read. In: CVPR. pp. 8317--8326
  (2019)

\end{thebibliography}

\end{document}